\documentclass[conference, letterpaper]{IEEEtran}
\ifCLASSINFOpdf
\else
\fi
\ifCLASSINFOpdf
   \usepackage[pdftex]{graphicx}
\else
\fi

\usepackage[cmex10]{amsmath}
\usepackage{color}
\usepackage{fancyhdr}
\usepackage[caption=false,font=footnotesize]{subfig}

\renewcommand{\thispagestyle}[2]{}

\fancypagestyle{plain}{
        \fancyhead{}
        \fancyhead[C]{first page center header}
        \fancyfoot{}
        \fancyfoot[C]{first page center footer}
}
\begin{document}

%
\title{MACHINE LEARNING APPLIED TO PERUVIAN VEGETABLES IMPORTS}

\author{\IEEEauthorblockN{Ticona-Salluca Hugo}
\IEEEauthorblockA{Faculty of Statistic and Computer Engineering,\\
Universidad Nacional del Altiplano de Puno, P.O. Box 291\\
Puno - Peru.\\
Email: hticonas@est.unap.edu.pe}
\and
\IEEEauthorblockN{Torres-Cruz Fred}
\IEEEauthorblockA{Faculty of Statistic and Computer Engineering,\\
Universidad Nacional del Altiplano de Puno, P.O. Box 291\\
Puno - Peru.\\
Email: ftorres@.unap.edu.pe}
\and
\IEEEauthorblockN{Tumi-Figueroa Ernesto Nayer }
\IEEEauthorblockA{Faculty of Statistic and Computer Engineering,\\
Universidad Nacional del Altiplano de Puno, P.O. Box 291\\
Puno - Peru.\\
Email: nayer.tumi@unap.edu.pe}}


%


\maketitle

\begin{abstract}
The current research work is being developed as a training and evaluation object.
the performance of a predictive model to apply it to the imports of vegetable products into Peru using artificial intelligence algorithms, specifying for this study the Machine Learning models: LSTM and PROPHET.
The forecast is made with data from the monthly record of imports of vegetable products(in kilograms) from Peru, collected from the years 2021 to 2022. As part of applying the training methodology for automatic learning algorithms, the exploration and construction of an appropriate dataset according to the parameters of a Time Series. Subsequently, the model with better performance will be selected, evaluating the precision of the predicted values so that they account for sufficient reliability to consider it a useful resource in the forecast of imports in Peru.
\end{abstract}



%
\IEEEpeerreviewmaketitle

\section{Introduction}
The Machine Learning methodology is one of the most valuable technologies in the Industry 4.0, and advances in Machine Learning have provided significant benefits in strategic decision-making in organizations\cite{perez2022aprendizaje}.
In recent years, predictive models using Machine Learning algorithms have been implemented in real environments of different organizations, and as in this particular case that concerns us, it also operates in the agriculture industry and optimal results are expected from these technologies.\par
Likewise, Machine Learning technology has a high potential to optimize your processes and, consequently, make correct and anticipated decisions. For this reason, Learning Algorithms should also be considered as the next innovative direction to significantly improve the prediction of these use cases. \cite{SARSHAR}.

Therefore, this research will seek to demonstrate and encourage the use of Machine Learning applications in the field of agribusiness in Peru. Through the present work, the examination and modeling of real data will be developed in order to forecast future values with the least possible prediction error. In this sense, this applied research work covers the design, training, validation, and evaluation of 2 predictive models. The research models to be explored are LSTM and Prophet models, these being adequate and recommended to develop applied foresting, on the one hand, LSTM is recognized for its hyperparameters which have the ability to capture non-linear patterns using the grid search method\cite{abbasimehr2020optimized} and PROPHET for their modular regression with intuitive and adjustable interpretation parameters applied on Time Series\cite{taylor2018forecasting}.

.
\section{DATASET}

According to the purpose of the study, the official records of the export volume (in kilograms) of vegetables in Peru were used, collected from 2021 to 2022 in the National Open Data Platform,\cite{datosabiertos}, officially uploaded by the Ministry of Agrarian Development and Irrigation of Peru.
\par
\includegraphics[width=8cm]{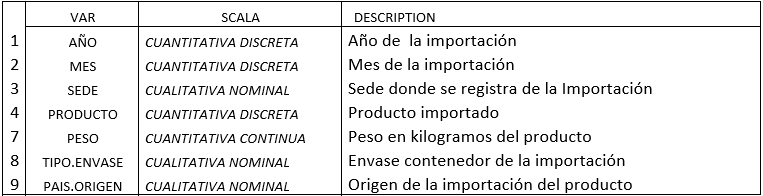}
Table 1\\
Description of the variables in the Vegetable Import Dataset in Peru 2021-2022.\par
Based on this dataset, it is proposed to make a reliable prediction, and according to the models to be developed, we will therefore need the continuous quantitative variable, PESO.\par 
Also according to the observed data, we find it necessary to adapt a proper timeline; therefore, we must join the variables AÑO and MES to foresee a more accurate seasonality, specifying the variables to forecast. Subsequently, to get a more accurate overview of our data regarding products, we define IDs for each product.
\par
\begin{figure}[!h] 
 \centering
 \includegraphics[width=.9\columnwidth]{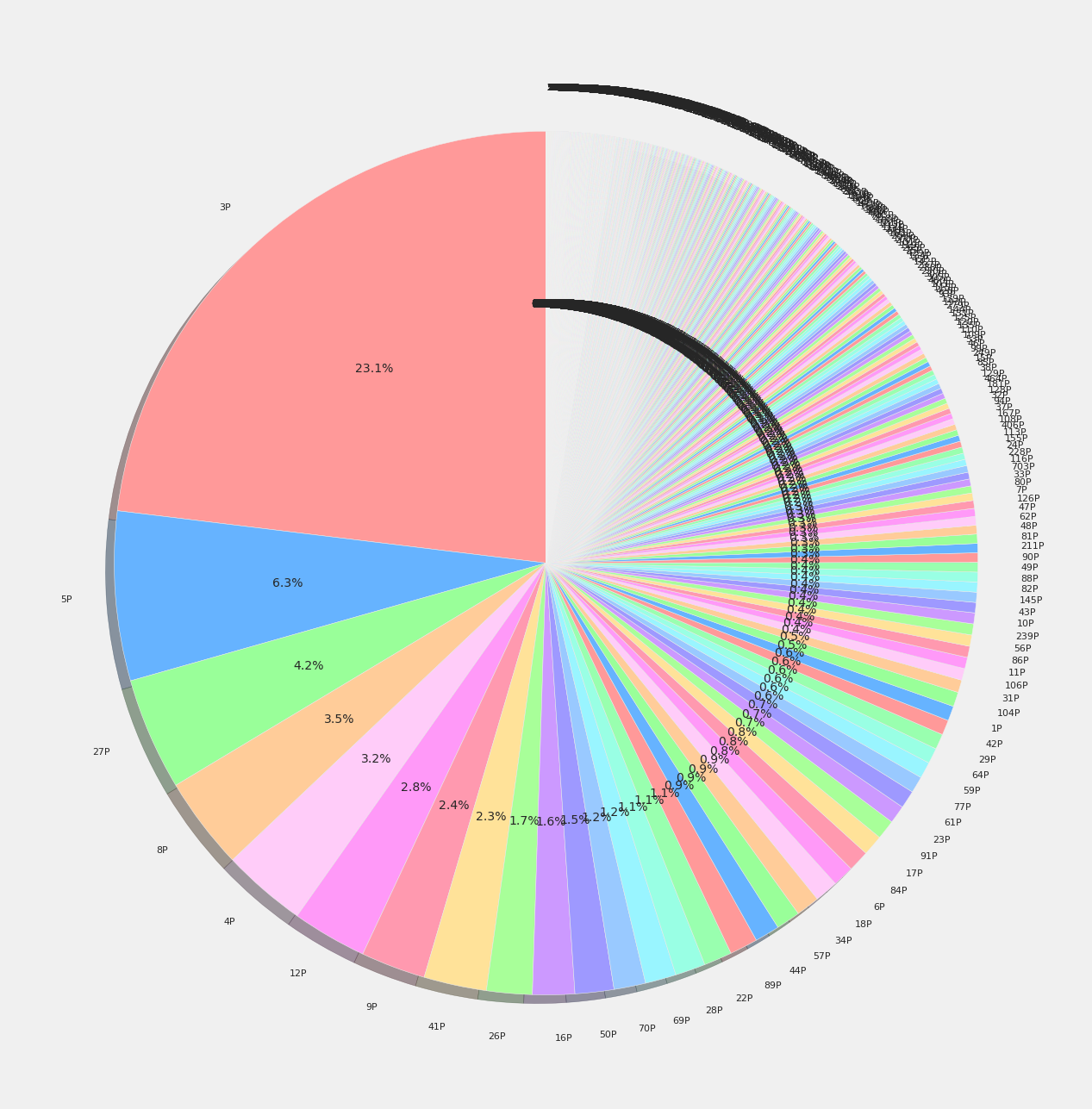} 
 \caption{Percentage description of each vegetable product imported in Peru} \label{fig-1}
\end{figure}
\par

From the dataset, we highlight 3P: Soja, cake with 23.1\% in 1st position, in second position 5P: Apple, fresh fruit with 6.3\% and in third position 27P: Soja, grain with 4.2\%. The data ranges from 2021-05-01 to 2022-06-01, with 34,364 observations on imports of a total of 848 plant products.


\section{METHODS}

In this section, we will describe the models used for the predictions.

\subsection{LSTM MODEL}

The LSTM name refers to the analogy that a standard RNN has both 'long-term memory' and 'short-term memory'. The connection weights and biases in the network change once per training episode, analogous to how physiological changes in synaptic strengths store long-term memories; activation patterns in the network change once per time step, analogous to how moment-to-moment changes in electrical activation patterns in the general brain store short-term memories. The LSTM architecture aims to provide a short-term memory for RNNs that can last for thousands of time steps and continue to be reliable, hence 'short-term long-term memory'\cite{elman1990finding}.

LSTM is considered a special type of recurrent neural network (RNN), developed to solve the potential problem of descending gradient found in traditional RNN training, and is able to learn both short-term and long-term dependencies\cite{sorkun2020time} and is constructed of four main components: an entry gate, an exit gate, memory cell and a forget gate.

Input Gate: controls the sending  addition of information to the cell state. In other words, the gateway will consider what information needs to be added to the cell state to ensure that important information is added and that there is no redundant information or noise.

Memory Cell: controls the value that might be deleted or updated, and contains a value that might need to be kept as additional information for many other time steps.

Output Gate: controls the selection of useful learning information from the current state of the cell as output.

Forget Gate: controls the removal of information that is no longer needed for LTSM to learn things or, less importantly, the state of the cell. This helps to optimize the performance of the LSTM proposed network.

Also, The LSTM model follows the sequence:

1. Decide to discard the cell state information by a Sigmoid layer called "forget gate".\par
2. Decide new information to store in the cell state. The Sigmoid layer called the "gateway layer" decides which values will be updated. the tan coat creates a vector of new candidate values that could be added to the state.\par
3. Update the old cell state to the new cell state.\par
4. Decide the filtered output based on the state of the cell.\par

\begin{figure}[!h]
	\centering
	\includegraphics[width=8cm]{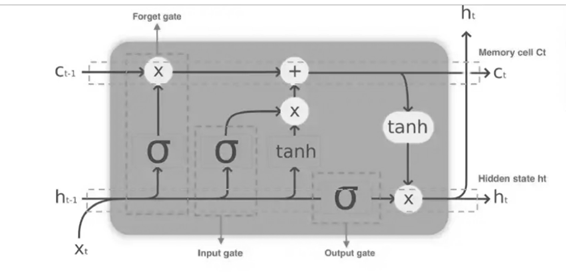}
	\caption{MODEL LSTM}
         
	\label{fig5}
\end{figure}
\par
Considered for a normal LSTM model is the Tanh property, which is a non-linear activation function. It regulates the values that flow through the network, keeping the values between -1 and 1. To avoid information fading, a function is needed whose second derivative can survive longer. There might be a case where some values become huge, causing the values to be irrelevant.\par
And of course an LSTM principal property: The Sigmoid that belongs to the family of nonlinear activation functions. Unlike Tanh, the Sigmoid holds values between 0 and 1. It helps the network update or forget data. If the multiplication results in 0, the information is considered forgotten. Similarly, the information is kept if the value is relevant\cite{liu2022use}.

This will assist the network in determining what data can be lost and what data should be kept.

\includegraphics[width=8cm]{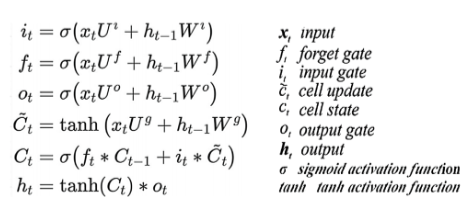}

\par

\subsection{PROPHET MODEL}
PROPHET is a procedure for forecasting time series data based on an additive model in which nonlinear trends are adjusted for annual, weekly, and daily seasonality. It works best with time series that have strong seasonal effects and multiple seasons of historical data. 

 In the specification of this Prophet model, there are several places where we can alter the model to apply your experience and external expert knowledge without needing to understand the underlying statistics.\cite{taylor2018forecasting}

For the Prophet model with general theory:\\
We use a mathematical decomposable time series model\cite{harvey1990estimation}
and this model has these components: trend, seasonality, and holidays. They are combined into the following equation:

\includegraphics[width=8cm\centering]{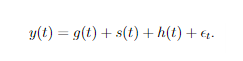}\\
Series Model Formula.

Here g(t) is the trend function that models non-periodic changes in the value of the time series, s(t) represents periodic changes (for example, weekly and yearly seasonality), and h(t) represents the effects of the seasons, The error term represents any idiosyncratic changes that do not fit the model. \par

In PROPHET we incorporate trend changes into the growth model by explicitly determining the change points where the growth rate is allowed to change. Suppose there are exchange points at moments j,j= 1,..., S. We give N a vector of adjustments, where only the rate change occurs at moment j. The rate at any time is then the base rate k, plus all adjustments up to the point of This is best represented by the vectors as follows:

\includegraphics[width=6cm\centering]{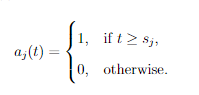}\\
Adjustments represented.

When rate k is adjusted, the parameter set must also be adjusted to connect the endpoints of the segments. The correct fit at the shift point is easily calculated as:

\includegraphics[width=8cm]{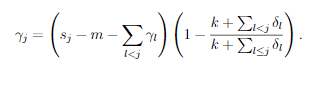}\par

That would be the Adjust at shift point.

The logistic piece of the growth model then looks as follows:
\par

\includegraphics[width=8cm]{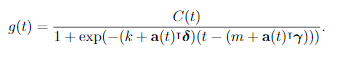}

The Fourier series is also applied in Prophet to provide a flexible model of periodic effects\cite{harvey199310}. Let P be the regular period that we expect the time series to have (for example, P= 365:25 for annual data or P= 7 for weekly data, when we scale our indices of time variables). We can approximate arbitrary uniform seasonal effects with this definition:

\includegraphics[width=7cm]{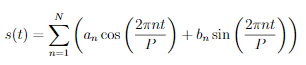}\\
Fourier Definition\par

Sometimes we can't just randomly split the data accordingly. PROPHET develops simulated historical forecasts by producing K forecasts at various cut-off points in history, chosen such that the horizons are within the historical record and the total error can be evaluated. \par
This procedure is based on the classic 'rolling source' forecast evaluation procedures\cite{tashman2000out}, but uses only a small sequence of target dates instead of forecasting by historical date) is that it saves on computation and provides less correlated precision measurements.

\subsection{Work Sequence}

\begin{itemize}
 \item We apply an EDA (Exploratory Data Analysis) it is necessary to clean the data and adapt it for the job.

 \item Normalization of the data, we adapt the data to follow a supervised sequence model according to a time series.
\item Let's divide our dataset into proof and validation tests.

 \item Coding and implementation of the models according to the Prophet or LSTM case.

  \item Adjustment of the parameters and extraction of predictions with their respective evaluation metrics.
\end{itemize}
\section{RESULTS}
The results using PROPHET and LSTM are shown in this section together with the general comparison of the mentioned models.\par
We will also appreciate the comparisons and metrics developed according to the models applied to the study variable: PESO.

Result according to the validation of the LSTM model:\par

\begin{figure}[!h]
	\centering
	\includegraphics[width=8cm,height=5cm]{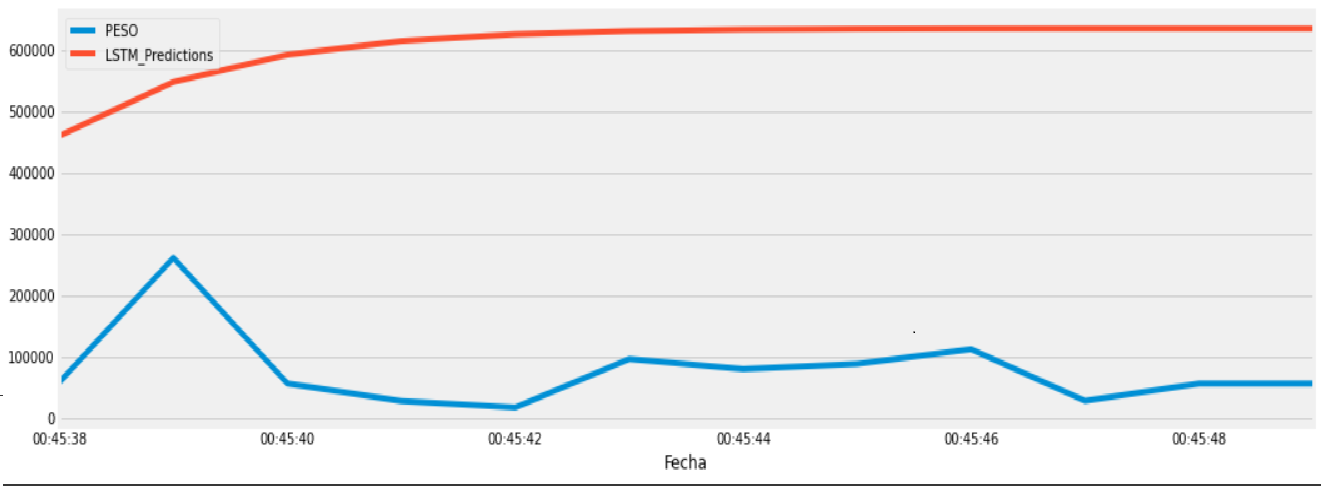}
	\caption{LSTM VALIDATION}
         
	\label{fig5}
\end{figure}
\clearpage
\par
Result according to the validation of the PROPHET model:\par

\begin{figure}[!h]
	\centering
	\includegraphics[width=8cm,height=5cm]{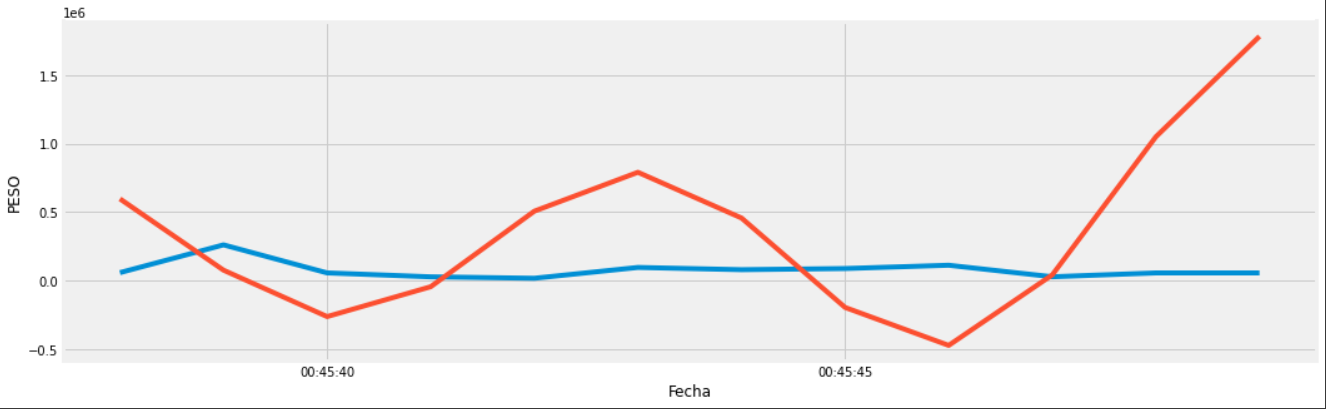}
	\caption{PROPHET VALIDATION}
         
	\label{fig5}
\end{figure}

Comparison to the validation of the PROPHET and LSTM model:\par

\begin{figure}[!h]
	\centering
	\includegraphics[width=8cm,height=5cm]{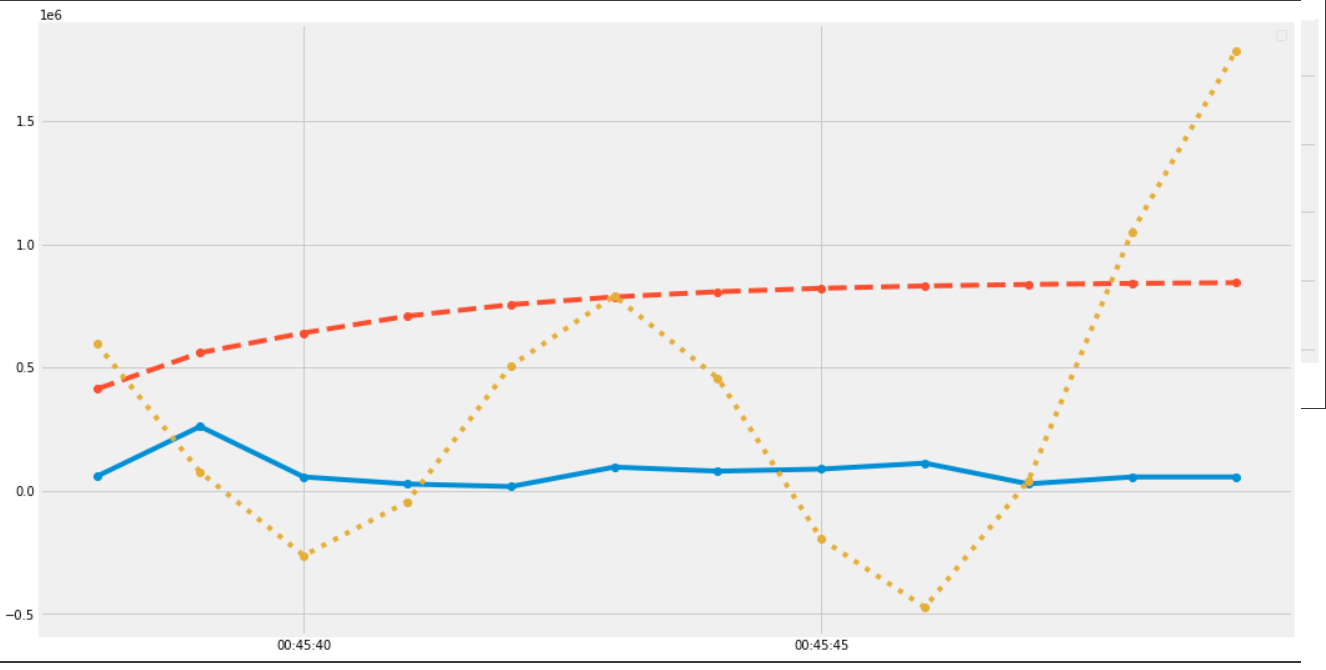}
	\caption{LSTM AND PROPHET COMPARISON}
         
	\label{fig5}
\end{figure}
\par

MSE(mean squared error) and the RMSE(root mean squared error) of both models:\par
We can appreciate each model according to its evaluation.\par
\begin{figure}[!h]
	\centering
	\includegraphics[width=8cm]{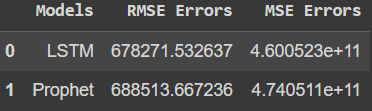}
	\caption{RMSE Y MSE }
         
	\label{fig5}
\end{figure}

\section{Discussion and Conclusions}
In keeping with the theory, all machine learning algorithms are unique, which is the root cause of why the prediction results are different algorithms on the same data set differ. LSTM was proposed for being an improvement of the RNNs and Prophet for its versatility in data with less presence of seasonality.\\
Given what was taken into consideration for LSTM,  the applied model was with the minimum parameters and show results with much better conditioning than was expected, remembering that in different studies the superiority of LSTM is detailed over the basic algorithms that apply Recurrent Neural Networks. Furthermore, it is noted from the theory that the number of training times, known as the 'epoch' in learning\cite{siami2018comparison}, did not take into account the expected effect on the performance of the trained forecast model and exhibited mostly random behavior. How intuitive The developed model based on LSTM incorporates additional 'gates' in order to store longer sequences of input data. \\
One of the main questions when developing and analyzing is whether the gates incorporated in the LSTM architecture would give a good prediction and if additional data training would be needed to further improve the prediction\cite{siami2019performance} and in this case, we can deduce that the quantity of data was acceptable but the predictions were affected by the not very well defined seasonality of the dataset. A more effective solution would be to add exact dates and continuous seasonalities.
\par
For the Prophet model, we should have some more intuitive results according to its theory, the application of the Fourier series could develop more precision. The model is expected to obtain a reasonable forecast on disordered data without too much manual effort, unlike LSTM, which has more hyperparameters. Prophet proved to be resistant to outliers, missing data, and drastic changes in its time series, the intention to fit the timeline is noted. Compared to other classical forecasting methods, Prophet should be fast and easy to apply to time series, which is what it was designed for in the first place; however, it could be considerably less accurate\cite{menculini2021comparing} and in this case, we confirm this appreciation by highlighting once again its intuitive factor.\\

The Prophet procedure should include more parameters for users to modify and adjust the forecasts in a more effective way. Also, a hybrid model could improve significantly. \\

According to the RMSE results of the import predictions, we can conclude that the LSTM model presents a significantly better performance and reliability with respect to the Prophet model, however, as we deduced previously, the seasonality of the dataset was an important key in the variation of the development of the models and their predictions. Therefore, increasing the size of the dataset and adapting an exact timeline for our dataset of vegetable imports from Peru should be a priority, in this way, we would undoubtedly obtain results with better relevance and reliability and, of course, the field of application with the use of machine learning techniques would be widely used and its results would be of a strongly necessary relevance.

\end{document}